\documentclass{article}

\PassOptionsToPackage{numbers, compress}{natbib}
\usepackage[final]{neurips_2019}

\usepackage[utf8]{inputenc} 
\usepackage[T1]{fontenc}    
\usepackage{hyperref}       
\usepackage{url}            
\usepackage{booktabs}       
\usepackage{amsfonts}       
\usepackage{nicefrac}       
\usepackage{microtype}      
\usepackage{multirow}
\usepackage{tablefootnote}
\usepackage{graphicx}
\usepackage{enumitem}
\usepackage{graphicx,array}
\usepackage{listings}
\usepackage[font=small,labelfont=bf,hypcap=false]{caption} 
\usepackage[newfloat,frozencache,cachedir=.]{minted}
\usepackage{standalone}
\usepackage{appendix}       
\usepackage{parcolumns}
\usepackage{xcolor}

\usepackage{inconsolata}
\usepackage[T1]{fontenc}
\title{PyTorch: An Imperative Style, High-Performance Deep Learning Library}

\author{
  Adam Paszke \\
  University of Warsaw \\
  \texttt{adam.paszke@gmail.com} \\
  \And
  Sam Gross \\
  Facebook AI Research \\
  \texttt{sgross@fb.com} \\
  \And
  Francisco Massa \\
  Facebook AI Research \\
  \texttt{fmassa@fb.com}
  \And
  Adam Lerer \\
  Facebook AI Research \\
  \texttt{alerer@fb.com}
  \And
  James Bradbury \\
  Google \\
  \texttt{jekbradbury@gmail.com}
  \And
  Gregory Chanan \\
  Facebook AI Research \\
  \texttt{gchanan@fb.com}
  \And
  Trevor Killeen \\
  Self Employed \\
  \texttt{killeent@cs.washington.edu}
  \And
  Zeming Lin \\
  Facebook AI Research \\
  \texttt{zlin@fb.com}
  \And
  Natalia Gimelshein \\
  NVIDIA \\
  \texttt{ngimelshein@nvidia.com}
  \And
  Luca Antiga \\
  Orobix \\
  \texttt{luca.antiga@orobix.com}
  \And
  Alban Desmaison \\
  Oxford University \\
  \texttt{alban@robots.ox.ac.uk}
  \And
  Andreas K\"{o}pf \\
  Xamla \\
  \texttt{andreas.koepf@xamla.com}
  \And
  Edward Yang \\
  Facebook AI Research \\
  \texttt{ezyang@fb.com}
  \And
  Zach DeVito \\
  Facebook AI Research \\
  \texttt{zdevito@cs.stanford.edu}
  \And
  Martin Raison \\
  Nabla\\
  \texttt{martinraison@gmail.com}
  \And
  Alykhan Tejani \\
  Twitter \\
  \texttt{atejani@twitter.com}
  \And
  Sasank Chilamkurthy \\
  Qure.ai \\
  \texttt{sasankchilamkurthy@gmail.com}
  \And
  Benoit Steiner \\
  Facebook AI Research \\
  \texttt{benoitsteiner@fb.com}
  \And
  Lu Fang \\
  Facebook \\
  \texttt{lufang@fb.com}
  \And
  Junjie Bai \\
  Facebook \\
  \texttt{jbai@fb.com}
  \And
  Soumith Chintala \\
  Facebook AI Research \\
  \texttt{soumith@gmail.com} \\
}

\begin{document}

\maketitle

\begin{abstract}
Deep learning frameworks have often focused on either usability or speed, but not both. PyTorch is a machine learning library that shows that these two goals are in fact compatible: it provides an imperative and Pythonic programming style that supports code as a model, makes debugging easy and is consistent with other popular scientific computing libraries, while remaining efficient and supporting hardware accelerators such as GPUs.

In this paper, we detail the principles that drove the implementation of PyTorch and how they are reflected in its architecture. We emphasize that every aspect of PyTorch is a regular Python program under the full control of its user. We also explain how the careful and pragmatic implementation of the key components of its runtime enables them to work together to achieve compelling performance.

We demonstrate the efficiency of individual subsystems, as well as the overall speed of PyTorch on several common benchmarks.

\end{abstract}

\section{Introduction} \label{introduction}

With the increased interest in deep learning in recent years, there has been an explosion of machine learning tools. Many popular frameworks such as Caffe~\cite{Caffe}, CNTK~\cite{CNTK}, TensorFlow~\cite{TF}, and Theano~\cite{Theano}, construct a static dataflow graph that represents the computation and which can then be applied repeatedly to batches of data. This approach provides visibility into the whole computation ahead of time, and can theoretically be leveraged to improve performance and scalability. However, it comes at the cost of ease of use, ease of debugging, and flexibility of the types of computation that can be represented. 

Prior work has recognized the value of dynamic eager execution for deep learning, and some recent frameworks implement this define-by-run approach, but do so either at the cost of performance (Chainer~\cite{Chainer}) or using a less expressive, faster language (Torch~\cite{Torch}, DyNet~\cite{DyNet}), which limits their applicability.

However, with careful implementation and design choices, dynamic eager execution can be achieved largely without sacrificing performance. This paper introduces PyTorch, a Python library that performs immediate execution of dynamic tensor computations with automatic differentiation and GPU acceleration, and does so while maintaining performance comparable to the fastest current libraries for deep learning. This combination has turned out to be very popular in the research community with, for instance, 296 ICLR 2019 submissions mentioning PyTorch.

\section{Background}

Four major trends in scientific computing have become increasingly important for deep learning.

First, starting in the 1960s, the development of domain specific languages such as APL \cite{APL}, MATLAB \cite{Matlab}, R \cite{R} and Julia \cite{Julia}, turned multidimensional arrays (often referred to as tensors) into first-class objects supported by a comprehensive set of mathematical primitives (or operators) to manipulate them. Separately, libraries such as NumPy\cite{Numpy}, Torch\cite{Torch}, Eigen\cite{eigenweb} and Lush\cite{Lush} made \textbf{array-based programming} productive in general purpose languages such as Python, Lisp, C++ and Lua.

Second, the development of \textbf{automatic differentiation}~\cite{autodiff_survey} made it possible to fully automate the daunting labor of computing derivatives. This made it significantly easier to experiment with different machine learning approaches while still allowing for efficient gradient based optimization. The autograd~\cite{maclaurin2016phd} package popularized the use of this technique for NumPy arrays, and similar approaches are used in frameworks such as Chainer~\cite{Chainer}, DyNet~\cite{DyNet}, Lush~\cite{Lush}, Torch~\cite{Torch}, Jax~\cite{jax} and Flux.jl~\cite{flux}.

Third, with the advent of the free software movement, the scientific community moved away from closed proprietary software such as Matlab\cite{Matlab}, and towards the \textbf{open-source Python ecosystem} with packages like NumPy \cite{Numpy}, SciPy \cite{SciPy}, and Pandas \cite{Pandas}. This fulfilled most of the numerical analysis needs of researchers while allowing them to take advantage of a vast repository of libraries to handle dataset preprocessing, statistical analysis, plotting, and more. Moreover, the openness, interoperability, and flexibility of free software fostered the development of vibrant communities that could quickly address new or changing needs by extending the existing functionality of a library or if needed by developing and releasing brand new ones. While there is a rich offering of open-source software for neural networks in languages other than Python, starting with Lush~\cite{Lush} in Lisp, Torch~\cite{Torch} in C++, Objective-C and Lua, EBLearn~\cite{EBLearn} in C++, Caffe~\cite{Caffe} in C++, the network effects of a large ecosystem such as Python made it an essential skill to jumpstart one's research. Hence, since 2014, most deep learning frameworks converged on a Python interface as an essential feature.
  
Finally, the availability and commoditization of general-purpose massively parallel hardware such as GPUs provided the computing power required by deep learning methods. Specialized libraries such as cuDNN~\cite{cudnn}, along with a body of academic work (such as \cite{maxdnn} and \cite{fast_cnn}), produced a set of high-performance reusable deep learning kernels that enabled frameworks such as Caffe~\cite{Caffe}, Torch7~\cite{Torch7}, or TensorFlow~\cite{TF} to take advantage of these \textbf{hardware accelerators}.

PyTorch builds on these trends by providing an array-based programming model accelerated by GPUs and differentiable via automatic differentiation integrated in the Python ecosystem.

\section{Design principles}

PyTorch's success stems from weaving previous ideas into a design that balances speed and ease of use. There are four main principles behind our choices:

\textbf{Be Pythonic} \quad
Data scientists are familiar with the Python language, its programming model, and its tools. PyTorch should be a first-class member of that ecosystem. It follows the commonly established design goals of keeping interfaces simple and consistent, ideally with one idiomatic way of doing things. It also integrates naturally with standard plotting, debugging, and data processing tools.

\textbf{Put researchers first} \quad PyTorch strives to make writing models, data loaders, and optimizers as easy and productive as possible. The complexity inherent to machine learning should be handled internally by the PyTorch library and hidden behind intuitive APIs free of  side-effects and unexpected performance cliffs.

\textbf{Provide pragmatic performance} \quad
To be useful, PyTorch needs to deliver compelling performance, although not at the expense of simplicity and ease of use. Trading 10\% of speed for a significantly simpler to use model is acceptable; 100\% is not. Therefore, its \emph{implementation} accepts added complexity in order to deliver that performance. Additionally, providing tools that allow researchers to manually control the execution of their code will empower them to find their own performance improvements independent of those that the library provides automatically.

\textbf{Worse is better}~\cite{worse_is_better} \quad
Given a fixed amount of engineering resources, and all else being equal, the time saved by keeping the internal implementation of PyTorch simple can be used to implement additional features, adapt to new situations, and keep up with the fast pace of progress in the field of AI. Therefore it is better to have a simple but slightly incomplete solution than a comprehensive but complex and hard to maintain design.

\section{Usability centric design}

\subsection{Deep learning models are just Python programs}
In a surprisingly short amount of time, machine learning grew from recognizing individual digits \cite{mnist} into autonomously playing StarCraft \cite{starcraft2}. Consequently, the neural networks themselves evolved rapidly from simple sequences of feed forward layers into incredibly varied numerical programs often composed of many loops and recursive functions.
To support this growing complexity, PyTorch foregoes the potential benefits of a graph-metaprogramming based approach to preserve the imperative programming model of Python. This design was pioneered for model authoring by Chainer\cite{Chainer} and Dynet\cite{DyNet}. PyTorch extends this to all aspects of deep learning workflows. Defining layers, composing models, loading data, running optimizers, and parallelizing the training process are all expressed using the familiar concepts developed for general purpose programming.

This solution ensures that any new potential neural network architecture can be easily implemented with PyTorch. For instance, layers (which in modern machine learning should really be understood as stateful functions with implicit parameters) are typically expressed as Python classes whose constructors create and initialize their parameters, and whose forward methods process an input activation. Similarly, models are usually represented as classes that compose individual layers, but let us state again that nothing forces the user to structure their code in that way. Listing \ref{lst:code_example} demonstrates how an entire model can be created by composing functionality provided by PyTorch such as 2d convolution, matrix multiplication, dropout, and softmax to classify gray-scale images. Note that linear layers are of course part of the library, but we show an example implementation to highlight how simple it is.

\begin{minipage}{\textwidth}
\begin{parcolumns}{2}
\colchunk{
\begin{minted}[fontsize=\small]{python}
class LinearLayer(Module):
   def __init__(self, in_sz, out_sz):
      super().__init__()
      t1 = torch.randn(in_sz, out_sz)
      self.w = nn.Parameter(t1)
      t2 = torch.randn(out_sz)
      self.b = nn.Parameter(t2)

   def forward(self, activations):
      t = torch.mm(activations, self.w)
      return t + self.b
\end{minted}
}
\colchunk{
\begin{minted}[fontsize=\small]{python}
class FullBasicModel(nn.Module):
   def __init__(self):
      super().__init__()
      self.conv = nn.Conv2d(1, 128, 3)
      self.fc = LinearLayer(128, 10)

   def forward(self, x):
      t1 = self.conv(x)
      t2 = nn.functional.relu(t1)
      t3 = self.fc(t1)
      return nn.functional.softmax(t3)
\end{minted}
}
\end{parcolumns}
\captionof{listing}{A custom layer used as a building block for a simple but complete neural network.}
\label{lst:code_example}
\end{minipage}
\bigskip

This ``everything is a just a program'' philosophy is not limited to just the models, and applies to optimizers and data loaders as well. This facilitates the experimentation of new training techniques. For example, to implement the very popular generative adversarial networks, one needs to specify two separate models (the generator and the discriminator), and two loss functions that depend on both models at the same time. Rigid APIs would struggle with this setup, but the simple design employed in PyTorch easily adapts to this setting as shown in Listing~\ref{lst:gan}.

\begin{figure}[thp]
\centering

\begin{minipage}{.6\textwidth}
\begin{minted}[fontsize=\small]{python}
discriminator = create_discriminator()
generator = create_generator()
optimD = optim.Adam(discriminator.parameters())
optimG = optim.Adam(generator.parameters())

def step(real_sample):
  # (1) Update Discriminator
  errD_real = loss(discriminator(real_sample), real_label)
  errD_real.backward()
  fake = generator(get_noise())
  errD_fake = loss(discriminator(fake.detach(), fake_label)
  errD_fake.backward()
  optimD.step()
  # (2) Update Generator
  errG = loss(discriminator(fake), real_label)
  errG.backward()
  optimG.step()

\end{minted}
\captionof{listing}{Simplified training of a generative adversarial networks.}
\label{lst:gan}
\end{minipage}
\end{figure}

Since PyTorch programs execute eagerly, all the features of Python are available throughout the whole design process. Print statements, standard debuggers, and common visualization tools like matplotlib all work as expected. Users do not have to wait for lengthy compilation before they can start running their programs, and more importantly intermediate computations can be observed to understand how a model works and whether its results are correct.

\subsection{Interoperability and extensibility}

Easy and efficient interoperability is one of the top priorities for PyTorch because it opens the possibility to leverage the rich ecosystem of Python libraries as part of user programs. Hence, PyTorch allows for bidirectional exchange of data with external libraries.
For example, it provides a mechanism to convert between NumPy arrays and PyTorch tensors using the \lstinline{torch.from_numpy()} function and \lstinline{.numpy()} tensor method. Similar functionality is also available to exchange data stored using the DLPack~\cite{dlpack} format.
Note that this exchange happens in both cases without any data copying -- objects on both sides only describe how to interpret a memory region which is shared among them.
Hence, those operations are actually extremely cheap, and take constant time no matter how large the converted arrays are.

Moreover, many of the critical systems are designed specifically to be extensible.
For instance, the automatic differentiation system allows users to add support for custom differentiable functions. To do that users can define a new subclass of \lstinline{torch.autograd.Function} that implements \lstinline{forward()} and \lstinline{backward()} methods, which specify the function and its derivative (or more formally the vector-Jacobian product).
Similarly new datasets can be added by subclassing \lstinline|torch.utils.data.Dataset|
and implementing two methods: \lstinline{__getitem__} (the indexing operator) and \lstinline{__len__} (the length operator), making datasets behave like (possibly lazy) lists. How these work is completely up to the implementer, and many users leverage other Python packages for data loading.
The \lstinline{DataLoader} class consumes objects conforming to this interface and provides an iterator over the data which takes care of shuffling, batching, parallelization, and management of pinned CUDA memory to improve throughput.

Most importantly, users are free to replace any component of PyTorch that does not meet the needs or performance requirements of their project. They are all designed to be completely interchangeable, and PyTorch  takes great care not to impose any particular solution.

\subsection{Automatic differentiation}

Since gradient based optimization is vital to deep learning, PyTorch must be able to automatically compute gradients of models specified by our users, and those can be arbitrary Python programs.
However, Python is a dynamic programming language that allows changing most behaviors at runtime, making ahead of time source-to-source differentiation cumbersome.
Instead, PyTorch uses the operator overloading approach, which builds up a representation of the computed function every time it is executed.
In its current implementation \cite{pytorch_autodiff}, PyTorch performs reverse-mode automatic differentiation, which computes the gradient of a scalar output with respect to a multivariate input. Differentiating functions with more outputs than inputs is more efficiently executed using forward-mode automatic differentiation, but this use case is less common for machine learning applications. PyTorch can be easily extended to perform forward-mode differentiation using array-level dual numbers \cite{Piponi-dual-numbers,Leuck-dual-numbers}.

Another interesting and uncommon feature of our system is that it can differentiate through code employing mutation on tensors, which is one of the basic building blocks of imperative programs.
To ensure safety, we have implemented a versioning system for tensors, which lets us track their modifications and ensure that we always use the data we expect.
One interesting tradeoff is that while we could utilize techniques like copy-on-write to support arbitrary programs, we chose to not go down this path, as performance-wise it is usually beneficial for the users to rewrite their code to ensure that no copies have to be performed.
Hence, while most mutations are benign and can be handled automatically, the really complicated cases result in a user error, which lets them know that they likely want to restructure the program.
This allows us to avoid introducing subtle and hard-to-find performance cliffs.

\section{Performance focused implementation}

Running deep learning algorithms efficiently from a Python interpreter is notoriously challenging: for instance, the global interpreter lock~\cite{python_gil} effectively ensures that only one of any number of concurrent threads is running at any given time. Deep learning frameworks based on the construction of a static data-flow graph sidestep this problem by deferring the evaluation of the computation to a custom interpreter.

PyTorch solved the problem differently, by carefully optimizing every aspect of its execution while simultaneously empowering its users to easily leverage additional optimization strategies.

\subsection{An efficient C++ core}

Despite being closely integrated in the Python ecosystem, most of PyTorch is written in C++ to achieve high performance. This core \lstinline{libtorch} library implements the tensor data structure, the GPU and CPU operators, and basic parallel primitives. It also provides the automatic differentiation system, including the gradient formulas for most built-in functions. This ensures that the computation of the derivatives of functions composed of core PyTorch operators is executed entirely in a multithreaded evaluator which does not require holding the Python global interpreter lock~\cite{python_gil}. Python bindings are generated using YAML meta-data files.
An interesting side-effect of this approach is that it allowed our community to quickly create bindings to multiple other languages resulting in projects like NimTorch \cite{nimtorch}, hasktorch \cite{hasktorch} and others.

This design also allowed us to create first-class C++ bindings and modeling libraries that can be used in places where Python is inconvenient, such as the game engine for Starcraft~\cite{starcraft_pytorch} or on mobile platforms. It is even possible to take the Python code describing a PyTorch model and run it without Python using the TorchScript engine~\cite{torchscript}. 

\subsection{Separate control and data flow}
PyTorch maintains a strict separation between its control (i.e. program branches, loops) and data flow (i.e. tensors and the operations performed on them). The resolution of the  control flow is handled by Python and optimized C++ code executed on the host CPU, and result in a linear sequence of operator invocations on the device. Operators can be run either on CPU or on GPU.

PyTorch is designed to execute operators asynchronously on GPU by leveraging the CUDA stream mechanism~\cite{cuda_stream} to queue CUDA kernel invocations to the GPUs hardware FIFO. This allows the system to overlap the execution of Python code on CPU with tensor operators on GPU. 
Because the tensor operations usually take a significant amount of time, this lets us saturate the GPU and reach peak performance even in an interpreted language with fairly high overhead like Python.
Note that this mechanism is nearly invisible to the user. Unless they implement their own multi-stream primitives all of the CPU-GPU synchronization is handled by the library.

PyTorch could leverage a similar mechanism to also execute operators asynchronously on the CPU. However the costs of cross-thread communication and synchronization would negate the performance benefit of such an optimization.

\subsection{Custom caching tensor allocator}

Almost every operator must dynamically allocate an output tensor to hold the result of its execution. It is therefore critical to optimize the speed of the dynamic memory allocators. PyTorch can rely on optimized libraries~\cite{hoard, jemalloc, tcmalloc} to handle this task on CPU. However, on GPU the \lstinline{cudaFree} routine may block its caller until all previously queued work on all GPUs completes. To avoid this bottleneck, PyTorch implements a custom allocator which incrementally builds up a cache of CUDA memory and reassigns it to later allocations without further use of CUDA APIs.
The incremental allocation is also crucial for better interoperability, because taking up all GPU memory ahead of time would prevent the user from utilizing other GPU-enabled Python packages.

To further improve its effectiveness, this allocator was tuned for the specific memory usage patterns of deep learning. For example, it rounds up allocations to multiples of 512 bytes to avoid fragmentation issues. Moreover, it maintains a distinct pool of memory for every CUDA stream (work queue).

The one-pool-per-stream design assumption simplifies the implementation and improves the performance of the allocator: because the CPU runs ahead of the GPU, memory is freed on the CPU \textit{before} its last use on the GPU finishes. Since streams serialize execution, if the free precedes the reallocation on the CPU, the same order will occur on the GPU. So the allocator can reallocate memory freed on the CPU immediately as long as the new allocation is used on the same stream as the freed region. However, if an allocation was last used on one stream and then allocated on another, additional synchronization is needed.

The one-pool-per-stream design seems limiting since the allocations end up fragmented per stream, but in practice PyTorch almost never uses multiple streams. It is notoriously hard to write CUDA kernels in a way that would let them cooperatively share the GPU because exact scheduling is hardware controlled. In practice, kernel writers usually resort to monolithic kernels that combine multiple tasks. Data loading and distributed computing utilities are exceptions to the one stream design, and they carefully insert additional synchronization to avoid bad interactions with the allocator.

While this design is susceptible to certain corner cases, it almost never exhibits unwanted behaviors in practical code. Most of our users are not aware of its existence.

\subsection{Multiprocessing}

Due to the global interpreter lock (GIL) Python's default implementation does not allow concurrent threads to execute in parallel.
To alleviate this problem, the Python community has established a standard \verb|multiprocessing| module, containing a number of utilities that allow users to easily spawn child processes and implement basic inter-process communication primitives.

However, the implementation of the primitives uses the same form of serialization used for on-disk persistence, which is inefficient when dealing with large arrays.
Hence, PyTorch extends the Python \verb|multiprocessing| module into  \verb|torch.multiprocessing|, which is a drop-in replacement for the built in package and automatically moves the data of tensors sent to other processes to shared memory instead of sending it over the communication channel.

This design greatly improves performance and makes the process isolation weaker, resulting in a programming model which more closely resembles regular threaded programs.
Users can easily implement heavily parallel programs that operate on independent GPUs but later synchronize gradients using all-reduce style primitives.

Another unique feature of this system is that it transparently handles sharing of CUDA tensors, making it easy to implement techniques like Hogwild \cite{Hogwild}.

\subsection{Reference counting}
Users often design their models to utilize all memory available during training, and increasing batch sizes is a common technique of speeding up the process. Therefore, to deliver great performance, PyTorch has to treat memory as a scarce resource that it needs to manage carefully.

Libraries with eager semantics have to manage tensor memory without knowing how it will be used in the future. Garbage collection is the typical way to handle this automatically because it has good amortized performance. In this approach, the runtime periodically investigates the state of the system, enumerates used objects and frees everything else. However, by deferring the deallocation, it causes the program to use more memory overall~\cite{garbage_collection}. Given the scarcity of GPU memory, these overheads are unacceptable. In fact, Torch7 utilized the garbage collector built into Lua, and a common anti-pattern among the users was to sprinkle the program with explicit triggers to the garbage collector, hoping that the memory errors go away.

PyTorch takes a different approach: it relies on a reference counting scheme to track the number of uses of each tensor, and frees the underlying memory \emph{immediately} once this count reaches zero. Note that PyTorch tracks both references internal to the \lstinline{libtorch} library and external references made by users in their Python code by integrating with Python's own reference counting mechanism. This ensures that memory is released exactly when tensors become unneeded.

One notable caveat is that we can only guarantee the desired performance characteristics in implementations of languages that either already utilize reference counting (CPython, Swift, but not PyPy or many scripting languages such as Lua), and those that allow for user-defined behavior for assignment, copies, and moves (e.g. C++, Rust).
Bindings to implementations that do not satisfy those criteria will have to implement their own specialized memory management on top of PyTorch.

\section{Evaluation}   \label{evaluation}

In this section we compare the performance of PyTorch with several other commonly-used deep learning libraries, and find that it achieves competitive performance across a range of tasks. All experiments were performed on a workstation with two Intel Xeon E5-2698 v4 CPUs and one NVIDIA Quadro GP100 GPU.

\subsection{Asynchronous dataflow}
We start by quantifying the ability of PyTorch to asynchronously execute dataflow on GPU. We use the built-in profiler \cite{autograd_profiler} to instrument various benchmarks and record a timeline of the execution of a single training step.

Figure \ref{fig:async_execution} shows a representative timeline of execution for the first few operations of a ResNet-50 model. The host CPU which queues the work quickly outpaces the execution of the operators on the GPU. This allows PyTorch to achieve almost perfect device utilization. In this example, GPU execution takes around three times longer than CPU scheduling. The exact ratio depends on the relative performance of the host CPU and the GPU, as well as the number of elements in each tensor and the average arithmetic complexity of the floating point computations to be performed on the GPU.

\begin{center}
  \includegraphics[width=\textwidth]{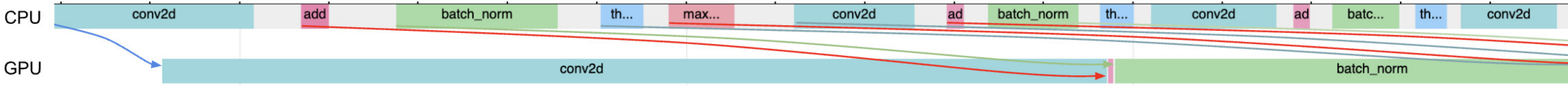}
  \captionof{figure}{A trace of the first few operators of Resnet-50. The top row depicts the execution of the control flow running on the host CPU. The gray areas are Python code executed by its interpreter. The colored areas correspond to the work done on the host CPU to queue various operators (convolution, batch normalization, and so on). The bottom row shows the corresponding execution of those operators on the GPU. The arrows pair the two events in time.\label{fig:async_execution}}
\end{center}

\subsection{Memory management}
We used the NVIDIA profiler to trace the execution of the CUDA runtime as well as the execution of the CUDA kernels launched during one training iteration of the ResNet-50 model. As shown in Figure~\ref{fig:resnet_annotated_traces}, the behavior of the first iteration differs significantly from that of subsequent ones. At first, calls to the CUDA memory management functions (\verb|cudaMalloc| and \verb|cudaFree|) slow down the execution quite dramatically by blocking the CPU thread for long periods of time, hence lowering the utilization of the GPU. This effect disappears in subsequent iterations as the PyTorch caching memory allocator starts reusing previously allocated regions.

\begin{center}
  \includegraphics[width=\textwidth]{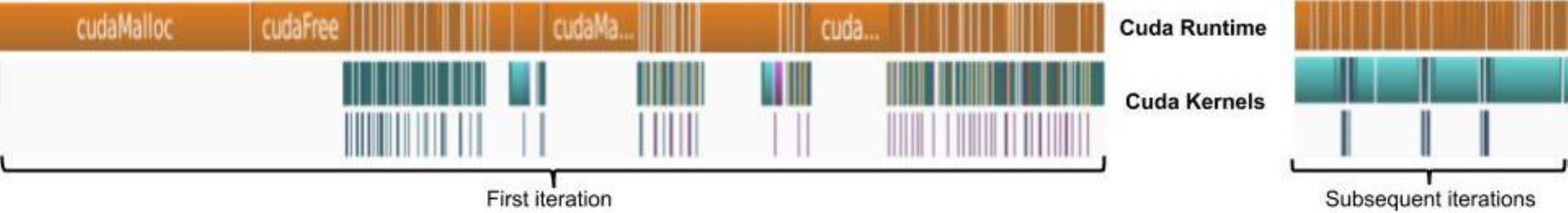}
  \captionof{figure}{Annotated traces of the execution of ResNet-50 on GPU. \label{fig:resnet_annotated_traces}}
\end{center}

\subsection{Benchmarks}

Finally, we can get an overall sense of single-machine eager mode performance of PyTorch by comparing it to three popular graph-based deep learning frameworks (CNTK, MXNet and TensorFlow), a define-by-run framework (Chainer), and production oriented platform (PaddlePaddle). The Appendix details all the steps needed to reproduce our setup.

Our results are summarized in Table~\ref{detailed_perf_results}. On all the benchmarks, the performance of PyTorch is within 17\% of that of of the fastest framework. We attribute this result to the fact that these tools offload most of the computation to the same version of the cuDNN and cuBLAS libraries.
\begin{table}[ht!]
\small
\centering
\begin{tabular}{l|c c c c c c}
 \toprule
 \multirow{2}{4.6em}{Framework} & \multicolumn{6}{c}{\it Throughput (higher is better)} \\
  & AlexNet & VGG-19 & ResNet-50 & MobileNet & GNMTv2 & NCF \\
\midrule

Chainer & $778 \pm 15$ & N/A & $\textbf{219} \pm 1$ & N/A & N/A & N/A \\
 CNTK & $845 \pm{8} $ & $ 84 \pm{3} $ & $210 \pm{1}$ & N/A & N/A & N/A \\
 MXNet & $\textbf{1554} \pm 22$ & $113 \pm 1$ & $218 \pm 2$ & $444 \pm 2$ & N/A & N/A \\
 PaddlePaddle  & $933\pm{123}$ & $112 \pm{2}$ & $192 \pm{4}$ & $\textbf{557}\pm{24}$ & N/A & N/A \\
 TensorFlow & $1422 \pm 27$ & $66 \pm 2$ & $200 \pm 1$ & $216 \pm 15 $ & $9631 \pm 1.3\%$ & $4.8e6 \pm 2.9\%$\\
 PyTorch & $1547 \pm 316$ & $\textbf{119} \pm 1$ & $212 \pm 2$ & $463 \pm 17$ & $\textbf{15512} \pm 4.8\%$ & $\textbf{5.4e6} \pm 3.4\%$\\

\bottomrule

\end{tabular}

 \caption{
  Training speed for 6 models using 32bit floats. Throughput is measured in images per second for the  AlexNet, VGG-19, ResNet-50, and MobileNet models, in tokens per second for the GNMTv2 model, and in samples per second for the NCF model. The fastest speed for each model is shown in bold.
  }
  \label{detailed_perf_results}

\end{table}

\subsection{Adoption}
The validity of design decisions and their impact on ease-of-use is hard to measure. As a proxy, we tried to quantify how well the machine learning community received PyTorch by counting how often various machine learning tools (including Caffe, Chainer, CNTK, Keras, MXNet, PyTorch, TensorFlow, and Theano) are mentioned on arXiv e-Prints since the initial release of PyTorch in January 2017. In Figure \ref{fig:pytorch_references} we report the monthly number of mentions of the word "PyTorch" as a percentage of all mentions among these deep learning frameworks. We counted tools mentioned multiple times in a given paper only once, and made the search case insensitive to account for various spellings.

\begin{center}
  \includegraphics[width=\linewidth]{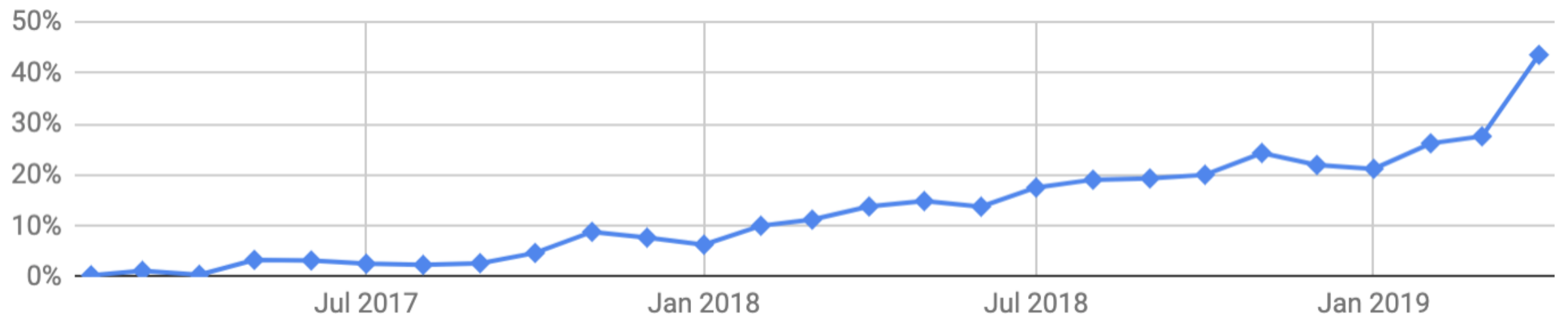}
  \captionof{figure}{Among arXiv papers each month that mention common deep learning frameworks, percentage of them that mention PyTorch.
  \label{fig:pytorch_references}}
\end{center}

\section{Conclusion and future work}
PyTorch has become a popular tool in the deep learning research community by combining a focus on usability with careful performance considerations. In addition to continuing to support the latest trends and advances in deep learning, in the future we plan to continue to improve the speed and scalability of PyTorch. Most notably, we are working on the PyTorch JIT: a suite of tools that allow PyTorch programs to be executed outside of the Python interpreter where they can be further optimized. We also intend to improve support for distributed computation by providing efficient primitives for data parallelism as well as a Pythonic library for model parallelism based around remote procedure calls.
\section{Acknowledgements}
We are grateful to the PyTorch community for their feedback and contributions that greatly influenced the design and implementation of PyTorch. We thank all the PyTorch core team members, contributors and package maintainers including Ailing Zhang, Alex Suhan, Alfredo Mendoza, Alican Bozkurt, Andrew Tulloch, Ansha Yu, Anthony Shoumikhin, Bram Wasti, Brian Vaughan, Christian Puhrsch, David Reiss, David Riazati, Davide Libenzi, Dmytro Dzhulgakov, Dwaraj Rajagopal, Edward Yang, Elias Ellison, Fritz Obermeyer, George Zhang, Hao Lu, Hong Xu, Hung Duong, Igor Fedan, Ilia Cherniavskii, Iurii Zdebskyi, Ivan Kobzarev, James Reed, Jeff Smith, Jerry Chen, Jerry Zhang, Jiakai Liu, Johannes M. Dieterich, Karl Ostmo, Lin Qiao, Martin Yuan, Michael Suo, Mike Ruberry, Mikhail Zolothukhin, Mingzhe Li, Neeraj Pradhan, Nick Korovaiko, Owen Anderson, Pavel Belevich, Peter Johnson, Pritam Damania, Raghuraman Krishnamoorthi, Richard Zou, Roy Li, Rui Zhu, Sebastian Messmer, Shen Li, Simon Wang, Supriya Rao, Tao Xu, Thomas Viehmann, Vincent Quenneville-Belair, Vishwak Srinivasan, Vitaly Fedyunin, Wanchao Liang, Wei Yang, Will Feng, Xiaomeng Yang, Xiaoqiang Zheng, Xintao Chen, Yangqing Jia, Yanli Zhao, Yinghai Lu and Zafar Takhirov.

\bibliographystyle{unsrt}
\bibliography{references}


\end{document}